\documentclass[preprint,12pt]{elsarticle}
\usepackage{color,colortbl}
\usepackage{graphicx}
\usepackage{amsmath}
\usepackage{url}
\usepackage{listings}
\usepackage{rotating}
\usepackage{changepage}
\usepackage{amssymb}
\begin{document}

% Define document data
% Option 1 for title:
\title{Identifying Table Structure in Documents using Conditional Generative Adversarial Networks
}

\author[a]{Nataliya Le Vine}
\author[b]{Claus Horn}
\author[c]{Matthew Zeigenfuse}
\author[d]{Mark Rowan}
\address[a]{Swiss Re, Digital and Smart Analytics, Armonk, New York, USA}
\address[b]{Swiss Re, Digital and Smart Analytics, Z\"urich, Switzerland}
\address[c]{Quartet Health, New York, New York, USA}
\address[d]{OTO Systems, Z\"urich, Switzerland \\
\rm{Email:} \textit{\{nataliya.levine,claus.horn,mdzeig\}@gmail.com, mark@tamias.co.uk }
}

% For shaded table rows
\definecolor{Gray}{gray}{0.9}

% Suppress name of References section
\renewcommand\refname{}

\begin{frontmatter}

\begin{abstract}
In many industries, as well as in academic research, information is primarily transmitted in the form of unstructured documents (this article, for example). Hierarchically-related data is rendered as tables, and extracting information from tables in such documents presents a significant challenge. Many existing methods take a bottom-up approach, first integrating lines into cells, then cells into rows or columns, and finally inferring a structure from the resulting 2-D layout. But such approaches neglect the available prior information relating to table structure, namely that the table is merely an arbitrary representation of a latent logical structure. We propose a top-down approach, first using a conditional generative adversarial network to map a table image into a standardised `skeleton' table form denoting approximate row and column borders without table content, then deriving latent table structure using xy-cut projection and Genetic Algorithm optimisation. The approach is easily adaptable to different table configurations and requires small data set sizes for training.
\end{abstract}


\end{frontmatter}

% Start the content
\section{Introduction}
\label{introduction}
\subsection{Problem outline}
In the last decades, humankind made many huge leaps forward in communications technology. We have witnessed the birth of the Internet; networks of computers allowing instantaneous digital transmission of knowledge. Yet, paradoxically, the primary method of information transmission in many industries and in academia, even in today's highly digital world, typically still takes the form of physical documents which are easy for humans to read but hard for computers to process. The text in these documents may now be digital characters in a PDF instead of ink printed on paper (although this is far from guaranteed, particularly in industries like insurance or in local government) but whether digital or printed, valuable hierarchical structure is typically discarded during the process of rendering a document.

Consider the common table: when hierarchically structured relational data is rendered as a table in a document, the data is converted into perhaps the most unstructured form imaginable (including in this article!) -- rendered as pixels of an image or characters on a page, where the only structure is to be found in the spatial location of pixels in relation to one another. The data is thus stripped of meta-information relating to its hierarchical structure, and information about relationships of columns and rows is lost.

Perhaps this comes as no surprise when one considers the Second Law of Thermodynamics: the entropy of the Universe always increases; order flows into chaos and structure is lost. It turns out that documents are another good example of this.

Common digital document exchange formats such as PDF do not explicitly represent any information about underlying data structures in tables. Instead, only individual characters and their locations on the page are encoded. Extracting and reimposing structure on data from tables in documents without reference to the original data structure from which they were created is a challenging problem in information extraction.

The idea to use a generative adversarial network (GAN) to extract table structures from scanned documents has been described in \cite{levine2019ijcnn}.
To make the approach applicable to real world applications a number of additional issues have to be addressed: 1) input images are usually skewed when manually scanned from documents, 2) a set of tables might have particular characteristics significantly different from other tables, 3) training might require a large training set, and 4) achieving stable training of GANs might be challenging. To alleviate this, the study explores the following areas:
training with blurry table structures, pre-processing input images to deskew them, model dependence on train set and its size, and, more generally, stable GAN training.

A major concern for applications of GANs is the stability of their training. Since the theoretical understanding of GAN behaviour is still limited, no automatic tuning procedure is available that would guarantee convergence to a stable equilibrium. Instead, we have to rely on a set of heuristics and best practices (see \ref{sec:GANtraining} for details).

\subsection{Mapping Manhattan: bottom-up vs top-down approaches}
\label{sec:bottom-up}
Existing approaches to the problem of extracting structured tabular information tend to be bottom-up. These approaches first identify pixel-level bounding boxes for table cells, exploiting regularities such as bounding lines and whitespace \cite{kasar2013learning,pinto2003table,gupta2019table}. The bounding boxes are subsequently aggregated into a table.

As a consequence, bottom-up approaches have a tendency to be brittle and sensitive to specific layout quirks. In some cases, they can also depend heavily on correctly setting tuning parameters, such as the number of white pixels required to infer a cell boundary.

The approach by Klampfl et al. \cite{klampfl2014comparison} uses hierarchical agglomerative clustering to group information at the word layout level, and subsequently builds up a table from sets of words which are assumed to belong to a single column. As with the approaches described previously, this is a bottom-up approach which places the graphical representation of the layout as the highest priority.

Focusing on the graphical structure in this way disregards available information relating to the higher-level structure of the table -- it is analogous to creating a street map of Manhattan by walking around at street level and trying to integrate the various turns into a larger map.

A top-down approach, by contrast, assumes that a table in a document is merely an arbitrary graphical representation of an unseen latent data structure. Wang \cite{wang1996tabular} specifies a model in which the \textit{logical} and \textit{presentational} forms of a table are completely decoupled: given an underlying set of data and its hierarchical inter-relations (the logical table), the presentation of such a table in a document is totally arbitrary with regard to its formatting, style, rotation, column widths and row heights, etc.

For a specific graphical representation of a logical table in a document which we want to extract, we can attempt to fit arbitrary representations of candidate logical table structures onto the graphical representation seen in the document. By introducing a measure of distance between the representation of a candidate table structure and the representation seen on the page, a gradient can be followed towards an optimal solution.

In this way, such an approach starts from the assumption that there is an underlying logical data structure which must be recovered, and decouples the search for the correct logical structure from the specifics of the graphical presentation, whilst using the graphical presentation to guide the search. In our street-map analogy, this makes use of the prior knowledge that Manhattan largely takes the form of a grid. Armed with this prior knowledge, we can overlay various different shapes and sizes of grids on top of an aerial photograph of Manhattan, adjusting the grid until it provides a good match to the visual information in the photograph.

\subsection{Overview of proposed top-down approach}
\begin{figure}[h]
    \begin{adjustwidth}{2cm}{}
    \includegraphics[scale=1.0]{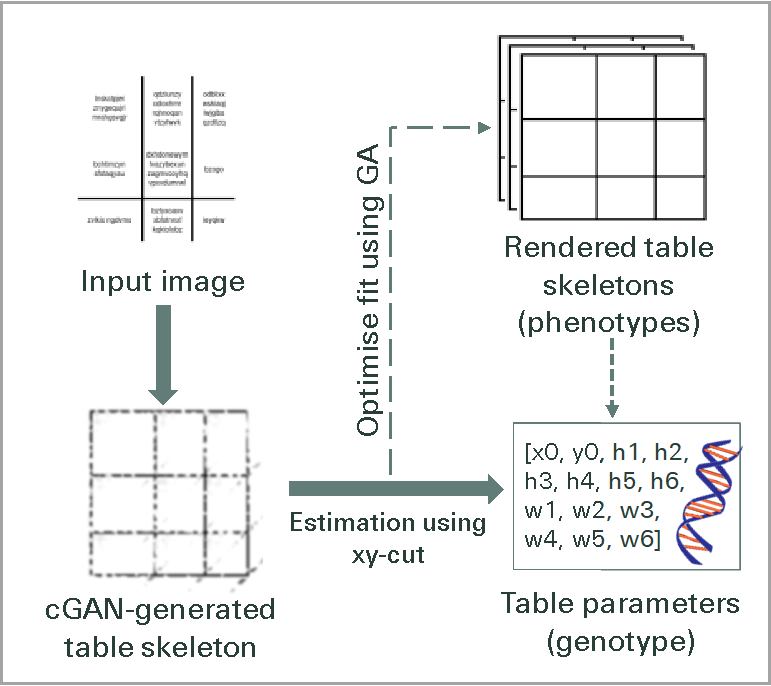}
    \end{adjustwidth}
    \caption{General schematic of the approach}
    \label{fig:skeleton}
\end{figure}

We propose a two-step generalisable top-down approach to extracting data from tables in documents (shown schematically in figure \ref{fig:skeleton}):
\paragraph{Step 1} Translate input images containing tables into an abstracted standard `skeleton' form showing pixel outlines for approximate locations of table cells and boundaries and disregarding actual table content.
\paragraph{Step 2} Produce a first table structure estimate using a simple projection method with the table skeleton from Step 1.
\paragraph{Step 3, optional} Optimise the fit of candidate latent data structures to the generated skeleton image using a measure of the distance between each candidate and the skeleton.

Once a good fit has been found, the data can be extracted from the table image and stored within the discovered structure using standard optical character recognition (OCR) techniques.

\subsection{Conditional GAN, xy-cut and Genetic Algorithm}
A conditional generative adversarial network (cGAN) adversarially trains two neural networks: a \textit{generator} learns to generate realistic ``fake'' candidates to fool a \textit{discriminator} which learns to detect the artificially generated input out of a set of real input examples. cGAN has shown remarkable success at generating realistic looking images from input data, for example transforming from aerial photographs to street maps, from greyscale to colour, or filling in outlines with realistic detail \cite{isola2017image}. We treat Step 1 of our table extraction method as another form of image translation, by training a cGAN to translate images containing tables into a standardised skeletons denoting table cell borders. Our method for training the cGAN is outlined in section \ref{sec:gan}.

An xy-cut method is a projection method that projects an image on x- and y-axis and analyses statistics of the derived projection distributions to approximate locations of horizontal and vertical lines on the original image \cite{nagy1984hierarchical}. More details about the method can be found in \ref{sec:initialguess}. 

Genetic algorithms (GA) use a population-based approach to sample the search space of possible solutions and to climb a gradient towards an optimum. The underlying representation of a candidate solution (dimensions and offsets of rows and columns of a table) is encoded as a vector of numbers denoted as the \textit{genotype}, which is decoded into a \textit{phenotype} (a rendered image of a table). Individuals within the population are then evaluated by a fitness function $f$ acting on the phenotypic representation. The GA optimisation is outlined in section \ref{sec:ga}.

\subsection{Improved Robustness of GAN training}
\label{sec:GANtraining}
Several papers have explored different possibilities to improve the stability of GAN training \cite{Salimans2016,Mescheder2018}.

We found that the following combination of approaches is helpful:
\begin{enumerate}
    \item Usage of batch normalization
    \item Usage of Gaussian weight initialization
    \item Usage of Adam for Stochastic Gradient Descent
    \item Usage of down-sampling (using strided convolutions) instead of pooling in the discriminator and up-sampling in the generator
\end{enumerate}

Moreover, working with gray-scale rather than black-and-white output images for training improves continuity and differentiability of the cGAN objective function. Further ideas, which could be added in the future, include adding Gaussian noise to discriminator input and use of additional training examples for the generator at each cycle.

\subsection{Comparison to other approaches}
A comprehensive comparison versus state-of-the-art systems is out of the scope of this paper, but nevertheless, to bring context to our approach we would like to comment on two representative commercial approaches: \textit{FlexiCapture} from Abbyy\footnote{http://help.abbyy.com/en-us/flexicapture/12/flexilayout\_studio/general\_tables} and \textit{CharGrid} from SAP \cite{katti2018chargrid}. \textit{FlexiCapture} is similar to other approaches we have seen, in that the user is required to manually define a hierarchical template for different ``classes'' of document (e.g. certain types of forms that the user regularly has to deal with), but the user is helped by some automatic estimation of table separators. This approach does not generalise well to previously unseen table layouts in new documents, unlike our approach which attempts to infer the latent data structure directly from the table image with no manual intervention.
\textit{CharGrid} \cite{katti2018chargrid} uses encoder-decoder convolutional neural networks to classify elements of a page and discover bounding boxes, including components of tables. This is in some ways similar to our approach for identifying table components visually in 2D space. In the case of tables, however, these elements must still be integrated together according to a bottom-up approach (see section \ref{sec:bottom-up}) based on character locations on the page. Whilst the \textit{CharGrid} representation for 2D text is powerful, it nevertheless does not make specific use of the prior assumptions relating to latent structure in table data which a top-down approach, such as we are proposing, can benefit from.

\section{Methods}
\label{method}
We formulate table structure as an image-to-genotype translation problem, whereby a table image is translated into a set of numbers (genotype) that fully characterise the table structure, i.e. column, row, and table positions and sizes. This is solved in the following stages: 1) translating an input table image (scan) into a corresponding table cell border outline image - table `skeleton', and 2) transforming the table skeleton into a latent table data structure - table genotype, and optionally 3) optimising the table genotype if needed.  (figure \ref{fig:skeleton}). 

The first stage masks out any text present in a table image, keeps any table cell borders present on the image, and adds any missing cell separators as implied by whitespace dividers in the input image. The tranformations are achieved using a conditional Generative Adversarial Network (cGAN) \cite{isola2017image}. Then, the second and third stages estimate and optimise (if needed) a table structure parameterisation using the table skeleton image derived at the first stage. The xy-cut projection method \cite{nagy1984hierarchical} estimates an initial table genotype. Meanwhile, Genetic Algorithm optimises table genotype in a way that the corresponding table phenotype (table rendered as an image) is close to the table skeleton produced by the cGAN (figure \ref{fig:skeleton}). The optimisation is performed within an a-priori selected class of tables, e.g. whether cell merging/splitting is allowed across rows/columns, or not. Note that, while it can be relaxed, an underlying assumption is that there is no text surrounding a table on an input image, and that all elements on an input image belong to one table.

\subsection{Table image to table skeleton transformation}
\label{sec:scan_to_skeleton}
A table image is mapped into a corresponding table skeleton (all row and column dividers, even if not present on the table image; and text masked out) using cGAN.

\subsubsection{cGAN architecture}
\label{sec:gan}
First, to reduce cGAN model complexity (number of parameters) and to optimise image-processing time, a table scan is resized to a smaller image of 256x256. Then the resized table image is transformed into a table skeleton using the cGAN architecture from \cite{isola2017image} as follows. 

A conditional GAN approximates a mapping from input image $x$ and random noise vector $z$ to output image $y$, so that $y=G(x,z)$. Generator $G$ is trained to produce outputs that cannot be distinguished from the target output images by a discriminator $D$ that is adversarially trained to detect `fake' images. Generator $G$ is an encoder-decoder network, so that an input image is passed through a series of progressively down-sampling layers until a bottleneck layer, where the process is reversed. To pass sufficient information to the decoding layers, a U-Net architecture with skip connections is used \cite{ronneberger2015u, isola2017image}, and a skip connection is added between each layer $i$ and $n-i$ via concatenation, where $n$ is the total number of layers, and $i$ is a layer number in the encoder \cite{isola2017image}. Further, following \cite{isola2017image}, random noise $z$ is provided only in the form of dropout applied on several layers of the generator at both train and test times.

The discriminator $D$ takes two images -- an input image and either a target (ground truth) or output image from the generator -- and assigns a probability that the second image is generated by a generator or not, i.e. the image is real. A convolutional PatchGAN architecture is used for the discriminator D \cite{li2016precomputed, isola2017image} that penalises output image structure at the scale of patches. In particular, the discriminator classifies whether each $N$ x $N$ patch in an image $y$ is real or fake, so that $D(x,y)$ equals $M$ x $M$ matrix containing probabilities for the patches in $y$ to be real, i.e. representative of a target image given input image $x$. We evaluated four discriminators  of varying complexity, so that each configuration has a different number of convolutional layers used ranging from three to six (for example, \cite{isola2017image} uses five layers for their image-to-image transformation).

\subsubsection{cGAN optimisation and inference}
The cGAN parameters are selected by optimising the following objective function with respect to generator $G$ and discriminator $D$:

\begin{equation}
	\label{eq:objective}
    \min_{G} \left[ \max_{D} L_{cGAN} (G,D) + \lambda L_{L1} (G) \right]
\end{equation}

Where $L_{L1}$ is $L1$ - distance between an image produced by a generator $G$ and an output image averaged over the training set, $\lambda$ is a constant weight ($\lambda=100$), and $L_{cGAN}$ expresses a cGAN objective to generate images that cannot be discriminated from real images:

\begin{equation}
    \begin{split}
    	\label{eq:lcgan}
        L_{cGAN} (G,D) &= E_{x,y} \left[\log D(x,y) \right]\\
                       &+ E_{x,y} \left[\log \left(1-D\left(x,G(x,z)\right)\right)\right]
    \end{split}
\end{equation}

Mixing of $L_{cGAN}$ and $L_1$ is found to be beneficial in previous studies, with $L_{cGAN}$ encouraging less blurring and $L_1$ suppressing artefacts \cite{pathak2016context, isola2017image}.

The cGAN is trained by alternating between optimising with respect to the discriminator $D$, and then with respect to the generator $G$ \cite{goodfellow2014generative}. Optimisation is performed on a set of 4,000 table scan/skeleton pairs (see section \ref{sec:traindata} for details of training set generation) using stochastic gradient descent applying the Adam solver \cite{kingma2014adam}, with learning rate 0.001, and momentum parameters $\beta_1$=0.9, $\beta_2$=0.999. Following \cite{ulyanov2016instance} and \cite{isola2017image}, the generator $G$ is used with dropout and `instance normalisation' (batch normalisation with a batch of 1) at inference time. 

\subsection{Random table generator}
\label{sec:traindata}
Due to the lack of empirical table image-to-skeleton and table image-to-genotype data available for training and testing, we synthetically generated the required sets using random table generator for four distinct table configurations.

\subsubsection{Table genotype}
\label{sec:genotype}
A table structure is described as a set of the following numbers, defined as table genotype:
\begin{itemize}
    \item table cardinality – number of rows $n$, and number of columns $m$;
    \item coordinates of the upper left corner of the table $\{x_0,y_0\}$;
    \item vector of row heights $\{h_1,\ldots,h_n\},h_i\geq0$;
    \item vector of column widths $\{w_1,\ldots,w_m\},w_i\geq$0.
\end{itemize}
The number or rows $n$ and number of columns $m$ given a-priori express an expected maximum table cardinality. For a specific table, when a number of rows and columns is smaller than $n$ and $m$, respectively, the corresponding row heights $h_i$ and column widths $w_i$ are set to zero in the table genotype. The genotype represents only a simple 2-D table class in the case when merging of table cells is disabled. The table class can be (but not pursued in this study) further extended by introducing a cell merge indicator array $\{c_{ij}\}_{(i,j=1)}^{(n,m)}$ into the table genotype, where $c_{ij}$ is set to a unique number indicating what type of cell merging (if any) is needed for cell in $i$th row and $j$th column.

\subsubsection{Table phenotype}
\label{sec:phenotype}

Each table genotype can be used to parameterise an XHTML representation of the table which is rendered into an image - table phenotype - using the imgkit Python library. Two types of tables borders are examined for the corresponding table skeletons - solid and blurry. The solid borders are parameterised in XHTML as:
\begin{lstlisting}[language=bash]
 border: 3px solid black; 
\end{lstlisting}
and the blurry borders are given by:
\begin{lstlisting}[language=bash]
 border: 3px solid black 
 box-shadow: inset 0px 0px 7px 3px black; 
\end{lstlisting}
so that the table border blur is 7 pixels, the spread is 3 pixels, and its offset is set to zero. The blurry borders allow accounting for the potential non-uniqueness of an optimal table skeleton corresponding to a table image, as reported in \cite{levine2019ijcnn}, where the authors note that moving a column or row border by a small number of pixels still results in a correct interpretation of a table structure. Furthermore, working with gray-scale rather than black-and-white images might help improving continuity of the cGAN objective function (eq. \ref{eq:objective}), especially, continuity of its $L_1$ component that accounts for distance between generated and output image. 

\begin{figure}[h]
    \includegraphics[width=\linewidth]{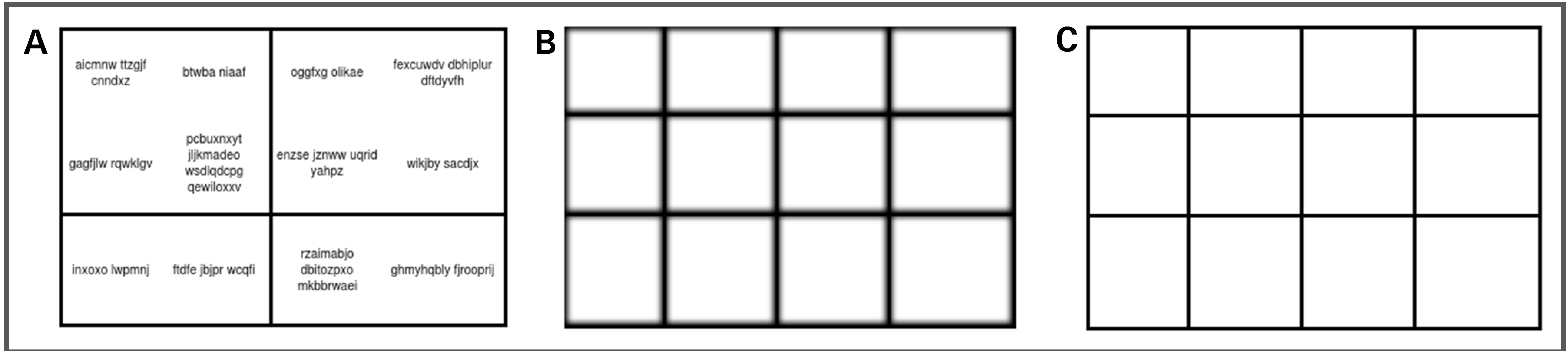}
    \caption{Examples of 'blurry and 'non-blurry' table pairs: A - table image, B - corresponding blurry table skeleton, and C - non-blurry table skeleton.}
    \label{fig:blurry_vs_nonblurry}
\end{figure}

Training and test sets for cGAN consist of table scan/skeleton image pairs, and a table scan image requires adding random text into the table cells as well as choosing, at random, row and column separators to be visualised in the rendered image. The corresponding table skeleton ground truth excludes the text and shows all column and row separators. Two sets of such table skeletons are generated - with non-blurry and blurry borders - figure \ref{fig:blurry_vs_nonblurry} panels A and B show an example of a blurry pair, and figure \ref{fig:blurry_vs_nonblurry} panels A and C show an example of a non-blurry pair.

Following \cite{levine2019ijcnn}, where it is noted that a cGAN model is sensitive to text font size, inter-column / inter-row spacing, and cell width and height, the model is trained on tables with four different configurations defined in table \ref{tab:tableconfigs}.  The training set and the test set each contain 4,000 tables, so that there are 1,000 tables of each of the four configurations.

\begin{table*} [h]
    \centering
    \caption{Four table configurations. ``$\cdot$'' refers to values the same as the `Base' configuration.}
    \label{tab:tableconfigs}
    \scriptsize
    \centering
    \begin{tabular}{|p{2.0cm}|p{0.6cm}p{0.6cm}p{0.9cm}p{0.9cm}p{0.9cm}p{1.1cm}p{0.9cm}p{0.9cm}p{0.9cm}|}
        \hline
        \multicolumn{1}{|c|}{} & \multicolumn{9}{c|}{}\\
        \textbf{Configuration} & \textbf{Rows} & \textbf{Cols} & \textbf{x-offset, ptx} & \textbf{y-offset, ptx} & \textbf{Row height, px} & \textbf{Column width, px} & \textbf{Word length, chars} & \textbf{Words per cell} & \textbf{Font size, px} \\
        \hline
        \multicolumn{1}{|c|}{} & \multicolumn{9}{c|}{}\\
        Base            & 2--6 & 2--6 & 0--70 & 0--70 & 40--90 & 70--100 & 5--9 & 2--4 & 10 \\
        
        Larger font   & $\cdot$ & $\cdot$ & $\cdot$ & $\cdot$ & $\cdot$ & $\cdot$ & $\cdot$ & $\cdot$ & 18 \\
    
        Smaller font    & $\cdot$ & $\cdot$ & $\cdot$ & $\cdot$ & $\cdot$ & $\cdot$ & $\cdot$ & $\cdot$ & 6 \\
        
        Short cells     & 4--10 & 4--10 & $\cdot$ & $\cdot$ & 20 & 40--60 & 1--4 & 1 & $\cdot$ \\
        \hline
    \end{tabular}    
    %\end{adjustwidth}
\end{table*}

\subsection{Estimating table structure for a table skeleton}
\label{sec:skeleton_to_struct}
The derived table skeleton is projected on horizontal and vertical axis to estimate the corresponding table structure; and, optionally, can be used to parameterise an objective function that defines table structure fitness for optimisation by Genetic Algorithm.

\subsubsection{Projection-based estimation of table structure}
\label{sec:initialguess}
Table structure is  estimated by projecting a colour-modified table skeleton onto horizontal and vertical axis (table borders are assumed nearly parallel to image borders) using the xy-cut method \cite{nagy1984hierarchical}. The colour modification classifies each image pixel as `black' or `white' using a threshold of 125 on pixel luminance value, so that pixels with luminance above the threshold are classified as `white', or `black' otherwise. The following RGB-to-luminance $E$ (``black-and-white'' portion of an image) transformation is used \cite{luminance}:

\begin{equation}
\label{eq:luma}
E=0.299 \cdot E_R + 0.587 \cdot E_G + 0.114 \cdot E_B
\end{equation} where $E_R$ is red intensity, $E_G$ is green intensity, and $E_B$ is blue intensity for a pixel; each intensity value ranges from 0 to 255.

It is noted that cGAN image may be somewhat noisy, so that an image may have a small number of short black line artifacts that do not correspond to any column/row separator. To minimise the effect, only those vertical/horizontal lines from the xy-cut are included that are at least 0.25 times as long as the longest identified vertical/horizontal line for the image. Both parameters -- the white-colour threshold (125) and the length parameter (0.25) -- are manually selected using an independent (validation) set of table images that is neither used for training nor testing. 

\subsubsection{Optimisation with genetic algorithm}
\label{sec:ga}
The initial table structure guess can be further optimised using a Genetic Algorithm (GA). GA uses reproduction, crossover, and mutation to evolve a population of tables from epoch $n-1$ to epoch $n$, so that table structure fitness is improved following the gradient of an objective function. Each table has its own genotype and phenotype as defined in section \ref{sec:traindata}, and the algorithm selects candidate tables based on the table fitness defined below. Parameters for GA, e.g. mutation rate, survival rate, are set to the values used in \cite{levine2019ijcnn} and are given below. The algorithm is specified to have converged when table fitness does not improve more than 1\% over three consecutive epochs.

In the algorithm, reproduction carries the best table structure over to the next epoch with no mutations (elitism), as well as 70\%  of other table structures with mutation. Further, the offspring mutation modifies table upper left corner coordinates, individual row heights, and column widths with a probability 0.1 for each entry in the genotype. The offspring mutation also modifies table structure (adding, merging, removing column/row) with probability of 0.1 per table dimension (columns or rows), so that the three structural operations are equally likely (probability of 0.03). Lastly, crossover is based on two parents, so that the upper left table x-coordinate and columns are inherited from the first parent, while the upper left table y-coordinate and rows are inherited from the second parent.

Following \cite{levine2019ijcnn}, the  objective function that we minimise is the fraction of non-overlapping non-white pixels between cGAN and the candidate image, calculated as:
	\begin{equation}
        \label{eq:min_overlap}
        \min_{u} \frac{|G(x,z)-u|_{L1}} {|1-u|_{L1} \cdot |1-G(x,z)|_{L1}}
    \end{equation} 
where pixels of the output image from cGAN $G(x,z)$ and a candidate table phenotype are scaled to values between 0 and 1, with 0 corresponding to a black pixel, and 1 corresponding to a white pixel. The objective function \ref{eq:min_overlap} varies between 0 and 1, so that the best match corresponds to 0 and the worst guess corresponds to 1.

As shown in \cite{levine2019ijcnn}, while being computationally intensive, a further table optimisation that follows an initial table estimation, brings limited improvements. The reason to include this procedure here is two-fold: 1) to assess GA benefits for tables with blurry borders (not considered in \cite{levine2019ijcnn}), and 2) to introduce an optimisation strategy for tables with more complex structures than the ones used in the work, e.g. tables with merged/split cells.

\subsection{Performance metrics}
\label{sec:perf}  

The proposed method is applied to images with 595x842 pixel resolution (corresponding to A4 paper size at 72 ppi) that are generated by the random table generator described in section \ref{sec:traindata} with parameters given in table \ref{tab:tableconfigs}. Conditional GAN is trained using 4,000 such table images, and a further 4,000 images are generated for performance evaluation, so that each set has 1,000 images for each of the four table configurations described in table table \ref{tab:tableconfigs}. The performance is assessed separately for each table configuration (table \ref{tab:tableconfigs}) by comparing row and column numbers, upper left corner positions, row heights and column widths and calculating the following metrics:

\begin{enumerate}
    \item Percentage of tables with correctly identified 
    \begin{itemize}
        \item row number,
        \item column number;
    \end{itemize}
    \item Average errors in incorrectly identified
    \begin{itemize}
        \item row number,
        \item column number;
    \end{itemize}
    \item Average errors in table upper left corner
    \begin{itemize}
        \item x-coordinate,
        \item y-coordinate;
    \end{itemize}
    \item average percentage errors in table
        \begin{itemize}
        \item row heights,
        \item column widths.
    \end{itemize}
\end{enumerate}

Errors are calculated as true value less corresponding predicted value, and relative errors are errors divided by corresponding true values. Average errors in column and row numbers are calculated only for tables where those are incorrectly identified. Meanwhile, error averages for column widths and row heights as well as average errors for upper left corner table coordinates are calculated only for tables with correctly identified numbers of columns and rows. This avoids double penalisation for identifying rows/columns incorrectly, e.g. when a top row is erroneously added or excluded, the y-coordinate is under-estimated or over-estimated, respectively. 

\subsection{Processing skewed table images}
\label{sec:skew}  
In order to asses the robustness with respect to skewed input images, additional test data sets were derived from the original test images by rotating each image by a fixed angle using the ImageMagic\cite{imgMagic} rotate command. Integer valued angles between -30 and +30 degrees were used in steps of 5 degrees resulting in 13 test sets. 

\begin{figure}[h]
    \begin{adjustwidth}{2cm}{}
     \includegraphics[scale=0.4]{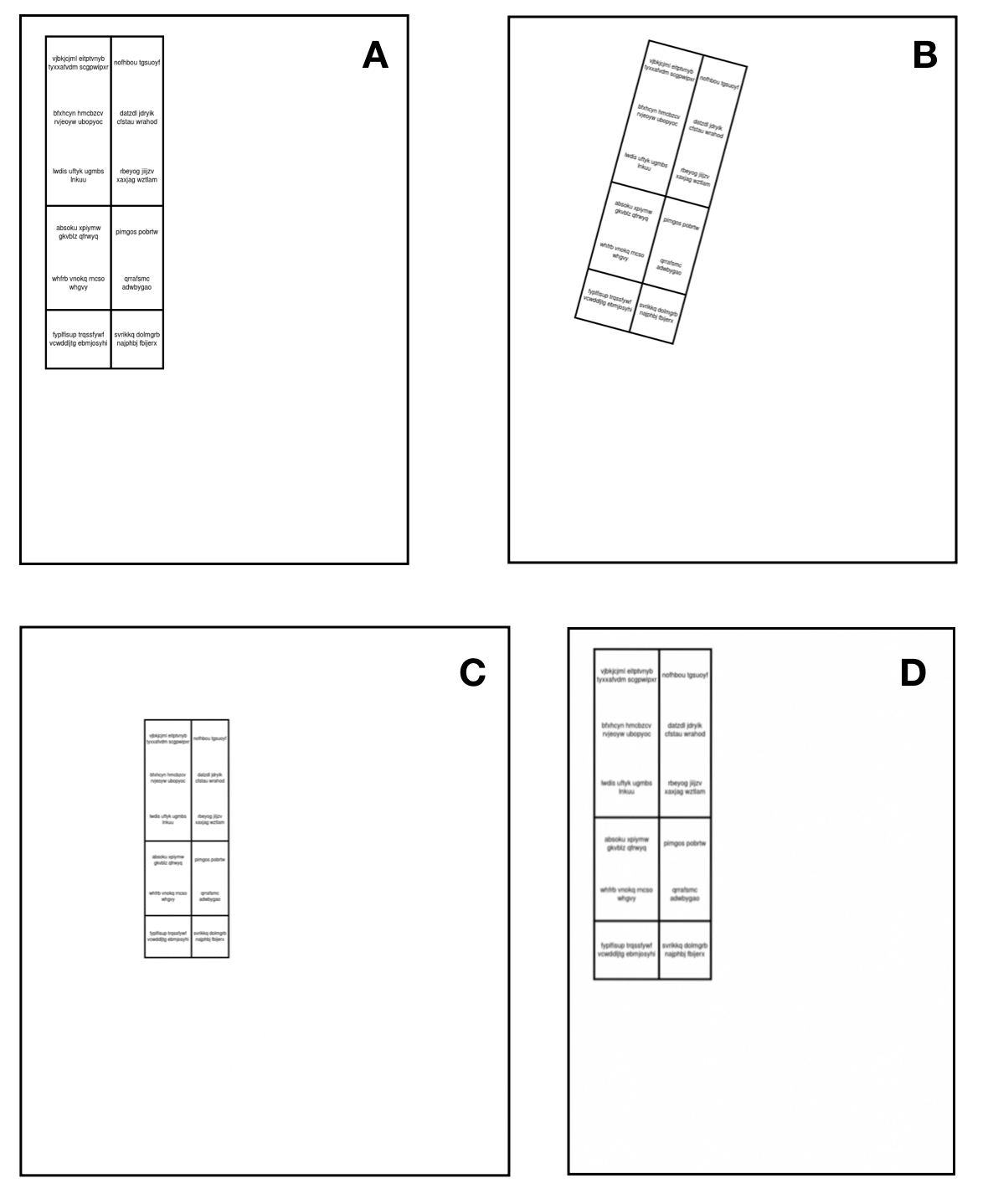}
    \end{adjustwidth}
    \caption{Illustration of the four-step  skewing/deskewing process. An example input table without skew (A). The same table rotated by 15 degrees (B). The result of applying the deskew algorithm with B as input (C). Cropping applied to C (D). The image sizes in pixels are: 595x842 (A), 795x970 (B), 1015x1140 (C), 595x842 (D).}
    \label{fig:rotationExperiment2}
\end{figure}

The deskewing process is shown for an example input table in figure \ref{fig:rotationExperiment2}. 
First the input images are assured to have skew angles compatible with 0, and then the different test sets are derived as described (the example of a 15-degree skew is shown in figure \ref{fig:rotationExperiment2}.) 
Images were rotated with the following command using the imageMagic package:
\begin{lstlisting}[language=bash]
 convert -background white -bordercolor white 
         -rotate D  A.png B.png
\end{lstlisting}
Where D is the positive or negative integer number of degrees. 
We changed the default from black to white background and border colour since black edges and triangles introduced by the rotation will otherwise confuse the GAN and reduce performance.  
% bash, Python

It is important to note, that both the rotation and deskewing (back rotation) operations translate images towards the bottom right as well as extend overall image dimensions (in number of pixeles) to accommodate the rotated image while keeping the original image content size constant within a larger image (Panel C of Fig \ref{fig:rotationExperiment2}). 

To satisfy the GAN input dimensions, the processed images were cropped as follows:
\begin{lstlisting}[language=bash]
 convert C.png -crop DXxDY+X0+Y0 D.png
\end{lstlisting}
Here DX and DY are dimensions of the output image (in pixels), and X0 and Y0 define the position of its upper left corner relative to the input (larger) image. The values were calculated by comparing the dimensions of images C and A (denoted $DX_C$ and $DX_A$ for the width and $DY_C$ and $DY_C$ for the height.):
\begin{equation}
        %\label{eq:fsa}
    \begin{array}{l}
        X0 = \frac{DX_C - DX_A}{2} \\ 
        Y0 = \frac{DY_C - DY_A}{2}
    \end{array}
\end{equation}
Their accuracy is thus directly related to the accuracy of the skew angle estimation. 

The deskewing is based on a Hough transformation using the deskew package\cite{deskew} as follows:
\begin{lstlisting}[language=bash]
 deskew -a 35 -b FFFFFF -o C.png B.png
\end{lstlisting} %\label{equ:deskew}
The maximal range of skew angles considered (using parameter $-a$ in the above command) is an important input, as performance decreases rapidly for angles larger than the default of $\pm 10$ degrees. 
Small remaining skew angles are found after the deskew transformation in many cases, especially for highly skewed input images. To improve this, the deskew command is applied multiple times in sequence (five times in this study), so that the remaining average skew angles are below 2 degrees for all input images.

\section{Experiments}
\label{sec:results}
We conducted the following types of experiments are conducted for table structure estimation to compare models trained using: 1) blurry vs. non-blurry table skeletons, 2) different complexities of the cGAN discriminator, 3) different sizes of train data sets. The xy-cut method is utilised to assess table structures, while the GA optimisation is only used to show improvements achievable when more time is available for table processing. Lastly, the algorithm is applied to example tables from ICDAR2013 competition \cite{icdar2013} to illustrate a need for a training set with more realistic table configurations.
 
First, a cGAN model with 5 layers in its discriminator is trained on 4,000 table pairs with blurry table skeletons (see figure \ref{fig:skeleton} A, B), and table structure is estimated using xy-cut method. The performance is compared with the model performance from \cite{levine2019ijcnn}, where an identical model configuration is used (cGAN with 5 layers in discriminator and xy-cut) that is also trained on 4,000 table pairs, but with non-blurry table skeletons (Table \ref{tab:results_blur}). 

The results reported in Table \ref{tab:results_blur} for the `non-blurry' model are somewhat different from the results reported in \cite{levine2019ijcnn}, due to slightly different metrics used as well as due to a different test set. When the `non-blurry' model is trained on the new data set generated for this paper with the same table configurations as in \cite{levine2019ijcnn}, the corresponding performance metrics are significantly different from the ones produced by the pre-trained model (based on three re-starts) -- the `base' and `larger font' configurations have significantly worse scores, and the `smaller font', and `short cells' configurations have better scores. 

Overall, table \ref{tab:results_blur} shows that the `blurry' model trained on blurry skeletons has a superior performance than the `non-blurry' model based on its high column and row number recall rates (97.9--100.0 for each table configuration), while the non-blurry recall rates are 64.4--96.6. 
The improvement might be due to the improved continuity and differentiability of the cGAN objective function (eq. \ref{eq:objective}), which operates with gray-scale, rather than black-and-white, output images when blurry model is trained.
Model error in column and row number is somewhat better for the non-blurry model, so that when the blurry model is wrong it makes a slightly larger error (misses/adds more rows/columns) than the non-blurry model. Note, the blurry table set with errors in column /row number is much smaller, so that the statistics are not directly comparable between the models. Lastly, all errors in the table upper left corner coordinates, column widths, row heights are small for both configurations, ranging between 0.2 and 5.0\% for the non-blurry model, and between 0.0 and 3.3. for the blurry model.   

\begin{table*}[h]
    \centering
    \caption{Metrics for models trained on blurry and non-blurry table skeletons with $5$ or $3$ convolutional layers in discriminator denoted as $D_5$, or $D_3$ evaluated on four table configurations - base (`Base' as shown in the table), smaller font (`Small'), larger font (`Large'), and short cells (`Short') configurations, all trained on $4,000$ table pairs. Best performances for correct number of rows and columns are given in bold font.}
    \label{tab:results_blur}
    \scriptsize
    \begin{adjustwidth}{-2.5cm}{}
    \begin{tabular}{|l|llll|llll|llll|}
        \hline
        % empty padding line
        \multicolumn{1}{|c|}{} & \multicolumn{4}{c|}{} & \multicolumn{4}{c|}{} & \multicolumn{4}{c|}{}\\
        % empty cell, then three range headers
        \multicolumn{1}{|c|}{} & \multicolumn{4}{c|}{\textbf{Non-blurry, $D_5$, $4,000$}} & \multicolumn{4}{c|}{\textbf{Blurry, $D_5$, $4,000$}} & \multicolumn{4}{c|}{\textbf{Blurry, $D_3$, $4,000$}} \\
        % Next row header
        \multicolumn{1}{|c|}{} & \multicolumn{4}{c|}{} & \multicolumn{4}{c|}{} &  \multicolumn{4}{c|}{} \\
        \textbf{Metric} & \textbf{Base} & \textbf{Small} & \textbf{Large} & \textbf{Short} &
                          \textbf{Base} & \textbf{Small} & \textbf{Large} & \textbf{Short} &
                          \textbf{Base} & \textbf{Small} & \textbf{Large} & \textbf{Short} \\
        \hline
        \multicolumn{1}{|c|}{} & \multicolumn{4}{c|}{} & \multicolumn{4}{c|}{} & \multicolumn{4}{c|}{}\\
        Correct row number, \%    & 88.7 & 90.8 & 64.4 & 81.1 & 98.6 & 98.1 & \textbf{97.9} & 98.7 & \textbf{99.8} & \textbf{98.9} & 96.4 & \textbf{99.4}\\
        Correct column number, \% & 92.6 & 92.5 & 91.6 & 96.6 & 99.6 & 98.1 & 93.3 & 100 & \textbf{99.7} & \textbf{99.3} & \textbf{99.3} & \textbf{100.0}\\
        Error in row number      & 0.6 & -0.3 & -1.0 & -0.6 & -1.0 & -1.2 & 1.0 & -1.0 & 0.0 & 0.3 & 1.1 & -0.7 \\
        Error in col. number &  -0.2 &  0.5 &  0.2 &  0.9 & -1.0 & -1.7 & 0.9 &   - &  -1.0 & -1.7 & 1.2 & -\\
        Error in x0, px &           1.0 &  1.7 &   -0.2 &  1.2 &  1.5 &  3.3 &  1.3 & 2.8 & 0.3 & 0.7 & -0.9 & 0.4\\
        Error in y0, px &          -1.5 & -0.1 &  -5.0 & -0.7 & -0.9 & -1.8 & -0.4 & -1.7 & -2.1 & -1.9 & -1.5 & -1.8 \\
        Error in column width, \% &  -0.3 & -1.5 &  0.0 & -0.9 &  0.5 & -0.1 & 0.8 & 0.1 & 1.1 & 1.5 & 1.7 & 1.1 \\
        Error in row height, \% &   -0.2 & -1.5 &  1.0  & -2.1 & 0.2  & 0.9  & 0.0 & 2.2 & 0.8 & 0.9 & 0.1 & 2.2\\
        \hline
    \end{tabular}
    \end{adjustwidth}
\end{table*}

Second, four other models are trained on 4,000 table pairs with blurry table skeletons, so that they differ by the number of convolutional layers in the cGAN discriminator -- six, four, and three layers (the original configuration adopted from \cite{isola2017image} has five layers). The six-layer configuration has much lower row/column number recall rates than the five-layer model for two of the table configurations -- `base' and `larger font', and similar performances for the other two table configurations (not shown). Meanwhile, the four- and three-layer models have, on average, slightly better recall rates, and the three-layer model has the best recall rates, except for its row number recall for the `larger font' configuration (Table \ref{tab:results_blur}).  

\begin{figure}[h]
    \center
    \includegraphics[scale=0.8]{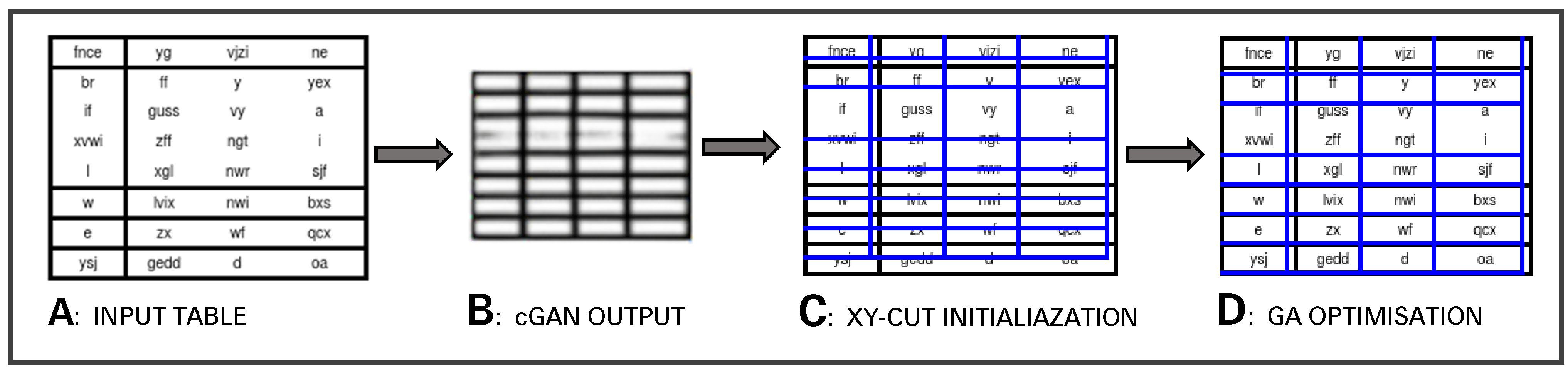}
    \caption{Procedure illustration example for a table image A: B - cGAN generated skeleton, C - xy-cut-based estimate (blue) overlaid with the table image, and D - GA-optimised estimate (blue) overlaid with table image.}
    \label{fig:pipeline_example}
\end{figure}

All performance metrics above are based on table structure estimated from a cGAN skeleton using the xy-cut method, and figure \ref{fig:pipeline_example} illustrates estimation procedure for a `short cells' table configuration that has the row number incorrectly identified (one row missed) as well misaligned row dividers (Figure \ref{fig:pipeline_example} C). To refine the estimate, table structure is further optimised using GA algorithm described in \ref{sec:ga}, so that the row dividers  align better, but the missed row is not added, due to the GA objective function specifics in equation \ref{eq:min_overlap}.

Third, the best performing `blurry' model with three layer discriminator trained on 4,000 samples is additionally trained on 10\% and 1\% of the data set size (400 and 40 tables, respectively) with equal number of tables from each table configurations (100 and 10 tables, respectively). Table \ref{tab:results_size} shows that the model trained only on 400 table pairs (10\% of the data) performs comparably to a model trained on 10 times the data (4,000 samples); and that the model trained on only 1\% of the original data size has high recall rates and low errors in table positioning and column/row sizes. Note, all models are assessed on the same test data set with 4,000 tables, so that each table configuration is represented by 1,000 tables.

\begin{table*}[h]
    \centering
    \caption{Metrics for models trained on different train set sizes - $4,000$, $400$, and $40$ table pairs with blurry skeletons, with $3$ convolutional layers in discriminator denoted as $D_3$, and evaluated on four table configurations -- base (`Base' as shown in the table), smaller font (`Small'), larger font (`Large'), and short cells (`Short') configurations. Best performances for correct number of rows and columns are given in bold font.}
    \label{tab:results_size}
    \scriptsize
    \begin{adjustwidth}{-2.5cm}{}
    \begin{tabular}{|l|llll|llll|llll|}
        \hline
        \multicolumn{1}{|c|}{} & \multicolumn{4}{c|}{} & \multicolumn{4}{c|}{} & \multicolumn{4}{c|}{}\\
         \multicolumn{1}{|c|}{} & \multicolumn{4}{c|}{\textbf{Blurry, $D_3$, $4,000$}} & \multicolumn{4}{c|}{\textbf{Blurry, $D_3$, $400$}} & \multicolumn{4}{c|}{\textbf{Blurry, $D_3$, $40$}} \\
         \multicolumn{1}{|c|}{} & \multicolumn{4}{c|}{} & \multicolumn{4}{c|}{} &  \multicolumn{4}{c|}{} \\
        \textbf{Metric} & \textbf{Base} & \textbf{Small} & \textbf{Large} & \textbf{Short} & 
                          \textbf{Base} & \textbf{Small} & \textbf{Large} & \textbf{Short} &
                          \textbf{Base} & \textbf{Small} & \textbf{Large} & \textbf{Short} \\
        \hline
        \multicolumn{1}{|c|}{} & \multicolumn{4}{c|}{} & \multicolumn{4}{c|}{} & \multicolumn{4}{c|}{}\\
        Correct row number, \%       &  \textbf{99.8} & \textbf{98.9} & 96.4 & \textbf{99.4} & 99.4 & 96.8 & \textbf{98.1} & 98.6 & 93.4 & 84.7 & 93.3 & 73.1\\
        Correct column number, \%     &  99.7 & \textbf{99.3} & 99.3 & \textbf{100.0} & \textbf{100.0} & 98.4 & \textbf{100.0} & \textbf{100.0} & 99.1 & 94.6 & 97.9 & 99.7\\
        Error in row number        &  0.0 & 0.3 & 1.1 & -0.7 &  -0.3 & -0.6 & -0.2 & -1.0 & 0.0 & -0.8 & -0.4 & -0.7\\
        Error in column number        & -1.0 & -1.7 & 1.3 & -  & - & -1.4 & - & - & -1.0 & -1.3 & -0.9 & -0.3\\
        Error in x0, px         &  0.3 & 0.7 & -2.0 & 0.4 & 0.5 & 0.6 & -0.4 & 0.3 & 2.0 & 1.7 & 1.9 & 0.5\\
        Error in y0, px         & -2.1 & -1.9 & -1.9 & -1.8 & -2.1 & -1.5 & -2.2 & -1.5 & -0.3 & 0.4 & -0.6 & -0.7\\
        Error in column width, \% &  1.1 & 1.5 & 1.7 & 1.1 & 1.3 & 1.9 & 1.0 & 1.2 & 0.0 & 1.0 & -0.2 & 0.8\\
        Error in row height, \% &  0.8 & 0.9 & 0.2 & 2.2 & 0.8 & 1.1 & 0.1 & 2.1 & -0.1 & 0.5 & -0.7 & 0.3\\
        \hline
    \end{tabular}
    \end{adjustwidth}
\end{table*}

\begin{figure}[h]
    \center
    \includegraphics[scale=0.85]{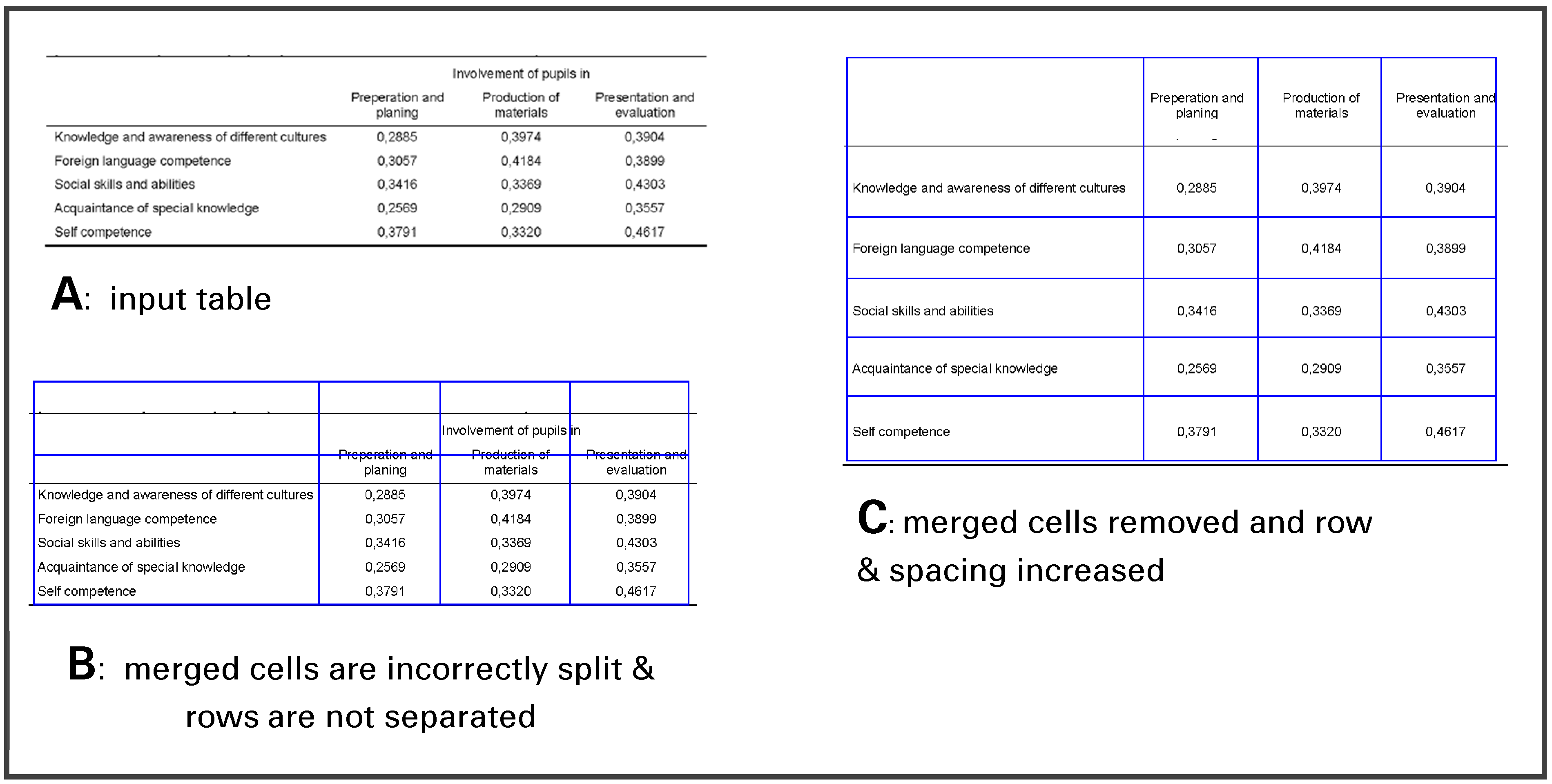}
    \caption{Table structure estimation example for a cropped out table A from ICDAR2013 table competition test set \cite{icdar2013}: B - estimate based on A as input table, C - estimate based on a modified input table A. The estimated structures are shown in blue.}
    \label{fig:icdar2013example}
\end{figure}

Lastly, the best model -- blurry model with three-layer discriminator, trained on 4,000 samples -- is applied to selected tables in the ICDAR2013 table competition data set \cite{icdar2013}. As the model is not trained to distinguish a table from document surrounding, e.g. other text and figures, test tables are cropped out (with some white buffers) from the  ICDAR2013 sample pages and used as input images to the model. Figure \ref{fig:icdar2013example} A illustrates a cropped table from the ICDAR2013 set, and shows that the model splits merged cells (as it is not designed to process such table configurations), and is unable to separate table rows due to the low inter-row spacing (Figure \ref{fig:icdar2013example} B). Removing the top merged cells as well as adding additional white space buffers between rows helps to improve the table structure estimate provided by the model (Figure \ref{fig:icdar2013example} C). 

Many tables in the ICDAR2013 table competition set have a much more complex structure than that shown in figure \ref{fig:icdar2013example} A and significantly deviate from the four considered table configurations, hence a full performance assessment is not attempted here, and only some identified model sensitivities are reported below:  

\begin{itemize}
    \item table location on the image,
    \item low or high intra-row and intra-column spacing,
    \item sensitivity to presence of empty cells.
\end{itemize}

Processing the ICDAR2013 tables will require training on more realistic generated tables or empirical tables (e.g. from ICDAR2013 train set) with corresponding paired table skeletons (which ICDAR2013 does not have)l . Additionally, it will require accounting for merged or split cells in the projection or/and optimisation algorithm steps.

\subsection{Robustness with respect to skewed input images}
\begin{figure}[h]
    \begin{adjustwidth}{-0.7cm}{}
        \centering\includegraphics[scale=0.4]{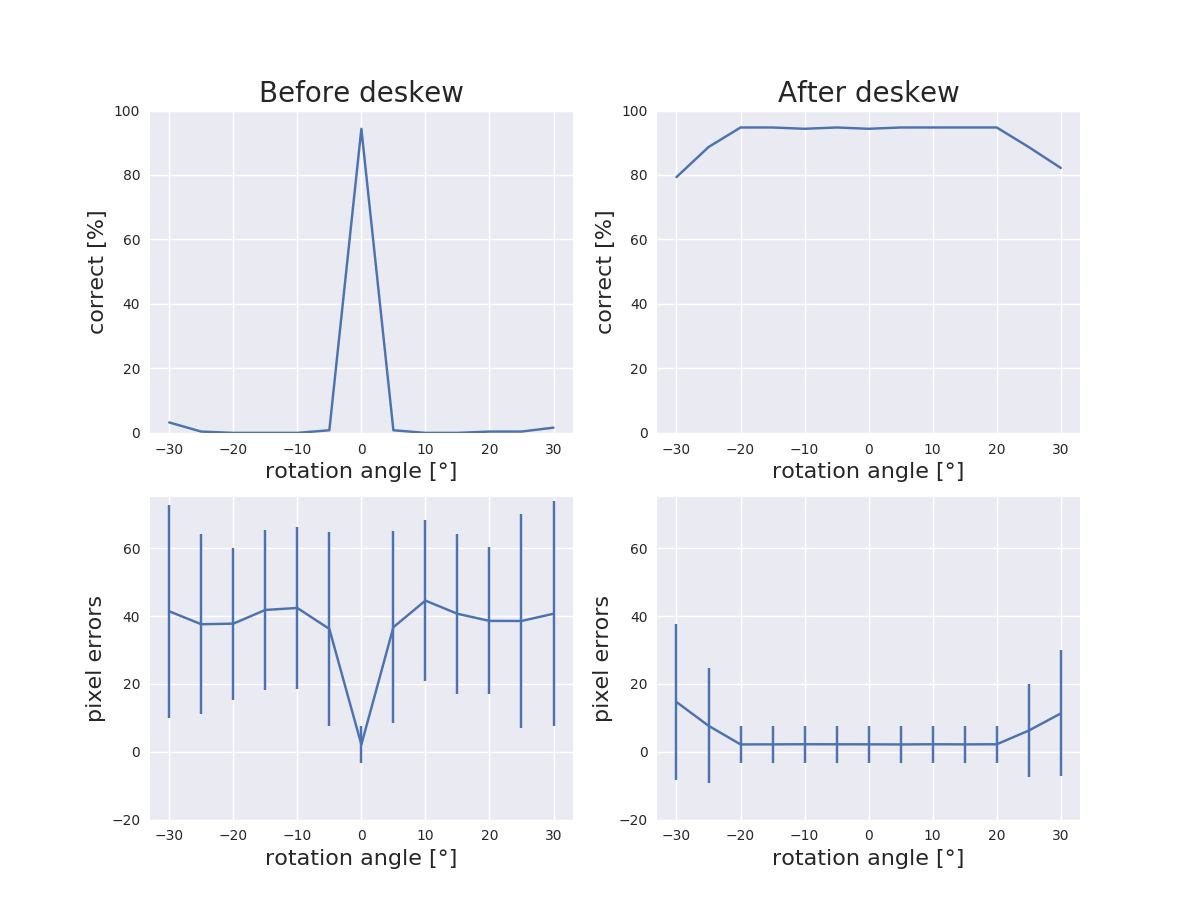}
    \end{adjustwidth}
    \caption{Performance of the table extraction algorithm as a function of rotation angle when rotated image is not corrected (left) and when the image is corrected with deskew pre-processing (right). The performance is measured by percentage of error-free table structures (top row), and by `pixel error' for row heights and column widths (bottom row).
    }
    \label{fig:rotationExperiment}
\end{figure}

Robustness with respect to skewed input images is an important aspect for the practical applicability of the table extraction technique. Figure \ref{fig:rotationExperiment} shows that the performance measured as percentage of error-free table structures quickly drops and reaches practically zero for absolute angle values greater than 5 degrees. 
Here, \textit{error-free table structures} means that both the number of rows and number of columns have been identified correctly. 
Furthermore, the \textit{pixel error} defined as a sum of average error in row heights (in pixels) and average error in column widths(in pixels)
increases to a maximum value for rotation angles different from 0, this maximum value corresponds to average cell width and height. 

Performing the deskew pre-processing as described in section \ref{sec:skew} allows recovery or nearly full performance for the range of tested input angles (see two right-hand-side panels of figure \ref{fig:rotationExperiment}). 
The decreasing precision of the deskew algorithm is reflected by small drops in performance for skewing angles above 20 degrees (and below -20 degrees) as observed in Fig. \ref{fig:rotationExperiment}. 

Lastly, a close inspection of the back-rotated image reveals that the lines loose sharpness, and hence might have slightly different characteristics than the tables used to train the model.

\section{Conclusions}
\label{sec:conclusions}
This study addresses the problem of estimating table structure from scanned table images. The problem arises in many applications working with paper documents or document scans, when structured layout has to be identified and converted into a database entry, or used for different information extraction tasks. The problem can have several valid solutions, as table row and column divider positions can be slightly perturbed and still accurately separate cell contents, due to padding around the dividers between rows and columns. For simplicity, the study focuses on a class of tables that does not contain merged or split cells; a further extension to a wider class of tables is theoretically possible, but left as an item for future research.

The problem is re-formulated as an image-to-genotype translation, and is solved in two stages: 1) translating an input table image into a corresponding table `skeleton' using cGAN, and 2) fitting a latent table structure (table genotype) to the derived table `skeleton' using the xy-cut method, and, optionally, Genetic Algorithm optimisation. When opted for, GA utilises a cGAN generated table structure skeleton for parameterising GA objective function that makes the function specific for each table image. 

The conditional GAN model is trained on pairs of table images and corresponding table structure `skeletons' where text is masked out and any missing row/column dividers are shown as blurry lines. This is different from the setup used in \cite{levine2019ijcnn}, when table structure skeletons are thin black lines of the same colour (`non-blurry' lines). The training set is generated by a random table generator and includes four distinct table configurations that differ by font size, column/row spacing, number of columns and rows, and cell content length. The derived solution has the following properties:
\begin{itemize}

    \item Using blurry column/row separators leads to more robust and stable model behaviour, as compared to the model trained on non-blurry table structure skeletons.
    
    \item Simpler cGAN structure outperforms more complex cGAN structures with a deeper discriminator network.
    
    \item The solution is adaptable to different table types, and table structure estimation quality is consistently high for all four considered table configurations.  

    \item Reducing the size of the 4,000-table training set to 10\% or 1\% of the original training set size (i.e. hundreds or dozens of tables) had no or limited effect on the prediction quality. 
    
    \item The cGAN model is sensitive to table and cell specifics, in particular large variations in row/column padding, word spacing, and table location on a page.
    
    \item Given the limitations of the xy-cut method as a projection-based method, tables with split/merged cells would require an optimisation method (such as GA) to further refine column and row dividers. 
    
    \item A pre-processing in form of deskewing allows high quality estimates when working with rotated table images, e.g. that are skewed due to document scanning.    
    
\end{itemize}

The approach appears promising, based on the high model adaptability to multiple table configurations that we demonstrate, and the low demands on training set size. To be used in practice, the method will need a training set with pairs of realistic tables and corresponding table structure skeletons. Furthermore, working with more complex table structure, e.g. with split/merge cells, would require additions to table genomes as well as a further optimisation of the xy-cut produced table estimates, for example by Genetic Algorithm. Lastly, the solution could be combined with an algorithm to detect the table area for tables in documents surrounded by text, as well as combined with an OCR algorithm to extract text into a data structure.

% Manually force references to next page to equalise the column lengths
\newpage

\section{References}

\bibliographystyle{elsarticle-num-names}
\bibliography{refs2}

\end{document}